\documentclass[letterpaper, 10 pt, conference]{ieeeconf}  

\IEEEoverridecommandlockouts                              

\usepackage{graphicx}
\usepackage{verbatim}
\usepackage{array}
\usepackage{upgreek}
\usepackage{float}   
\usepackage{amsmath,amssymb}
\usepackage{cite}
\usepackage{booktabs}
\usepackage{algorithm}
\usepackage[noend]{algpseudocode}
\usepackage{siunitx}

\usepackage{todonotes}
\usepackage{xcolor}

\usepackage[normalem]{ulem}

\makeatletter
\def\algbackskip{\hskip-\ALG@thistlm}
\makeatother

\usepackage{tikz}
\usetikzlibrary{positioning, shapes, arrows.meta}
\usetikzlibrary{shapes,arrows,chains}
\usetikzlibrary{arrows,calc,automata}

\tikzset{
    block/.style = {draw, rectangle, 
        minimum height=.5cm, 
        minimum width=1cm},
    input/.style = {coordinate,node distance=.5cm},
    output/.style = {coordinate,node distance=.5cm},
    arrow/.style={draw, -latex,node distance=.5cm},
    pinstyle/.style = {pin edge={latex-, black,node distance=1cm}},
    sum/.style = {draw, circle, node distance=1cm},
    line/.style={-{Stealth}}
    }

\title{\LARGE \bf
Empirical Study of Ceiling Proximity Effects and Electrostatic Adhesion for Small-scale Electroaerodynamic Thrusters
}

\author{
C. Luke Nelson$^{1}$, Grant Nations$^{2}$, and Daniel S. Drew$^{3}$ 
\thanks{$^{1}$Department of Mechanical Engineering, University of Utah, Salt Lake City, UT 84112, USA}
\thanks{$^{2}$Kahlert School of Computing, University of Utah, Salt Lake City, UT 84112, USA}
\thanks{$^{3}$Department of Electrical and Computer Engineering, University of Hawaii at Manoa, Honolulu, HI 96822, USA}
\thanks{Corresponding author: Daniel S. Drew, \tt{ddrew@hawaii.edu}}
}

\begin{document}

\maketitle
\thispagestyle{empty}
\pagestyle{empty}

\begin{abstract}

Electroaerodynamic propulsion, where force is produced via the momentum-transferring collisions between accelerated ions and neutral air molecules, is a promising alternative mechanism for flight at the micro air vehicle scale due to its silent and solid-state nature. 
Its relatively low efficiency, however, has thus far precluded its use in a power-autonomous vehicle; leveraging the efficiency benefits of operation close to a fixed surface is a potential solution. 
While proximity effects like the ground and ceiling effects have been well-investigated for rotorcraft and flapping wing micro air vehicles, they have not been for electroaerodynamically-propelled fliers.
In this work, we investigate the change in performance when centimeter-scale thrusters are operated close to a ``ceiling'' plane about the inlet. We show a surprising and, until now, unreported effect; a major electrostatic attractive component, analogous to electroadhesive pressure but instead mediated by a stable atmospheric plasma. 
The isolated electrostatic and fluid dynamic components of the ceiling effect are shown for different distances from the plane and for different materials. We further show that a flange attached to the inlet can vastly increase both components of force. A peak efficiency improvement of 600$\%$ is shown close to the ceiling. This work points the way towards effective use of the ceiling effect for power autonomous vehicles, extending flight duration, or as a perching mechanism.

\end{abstract}
\section{Introduction}
With continued advances in battery and microcontroller technology, micro air vehicles (MAVs) are poised to cross the gap from impractical to ubiquitous~\cite{xu2023high,mulgaonkar2014power,raza2021energy}. Small flying robots are well suited for reconnaissance, infrastructure inspection, and monitoring of agriculture, factories, and construction sites, both alone and in swarms~\cite{awasthi2020artificial,hafeez2023implementation,yoon2023cnn,elmokadem2021towards}. Many of these applications will be in situations where close operation to surfaces is necessary. In these cases, considering the impacts of surface proximity on flight performance is crucial for safety, precision, and to extend flight duration~\cite{ding2023aerodynamic,britcher2021use,yu2023adaptive}.

In many applications, it is beneficial for MAVs to deploy and then remain stationary at a specific location. Perching is an interesting way to increase operational lifetime, and it has been explored using a variety of mechanisms~\cite{liu2020adaptive,thomas2013avian,hsiao2023energy}. Among these mechanisms, electrostatic adhesion (``electroadhesion''), which works via a standing electric field inducing polarization and subsequent electrostatic attraction to nearby surfaces, is particularly advantageous for MAVs due to the increasing surface area-to-volume ratio at small scales. This makes it an effective solution for perching without the need for complicated mechanical designs or control systems~\cite{graule2016perching}. 

Rotorcraft and flapping wing MAVs have been the focus of research for decades and have seen significant progress towards real-world impact. Commercial MAV-scale rotorcraft offer high levels of out-of-the-box functionality, while flapping-wing MAVs have seen advancements in actuator performance and aerodynamic understanding which has led to dramatically enhanced flight efficiency~\cite{chen2019review}. In contrast, electroaerodynamic (EAD)-propelled robots represent a new class of MAVs and offer an interesting alternative due to their silent operation and, as solid-state devices, their mechanical simplicity. They are highly scalable and can be mass manufactured using silicon batch fabrication processes~\cite{drew2017first}.

\begin{figure}[t]
    \centering
    \includegraphics[width=\columnwidth]{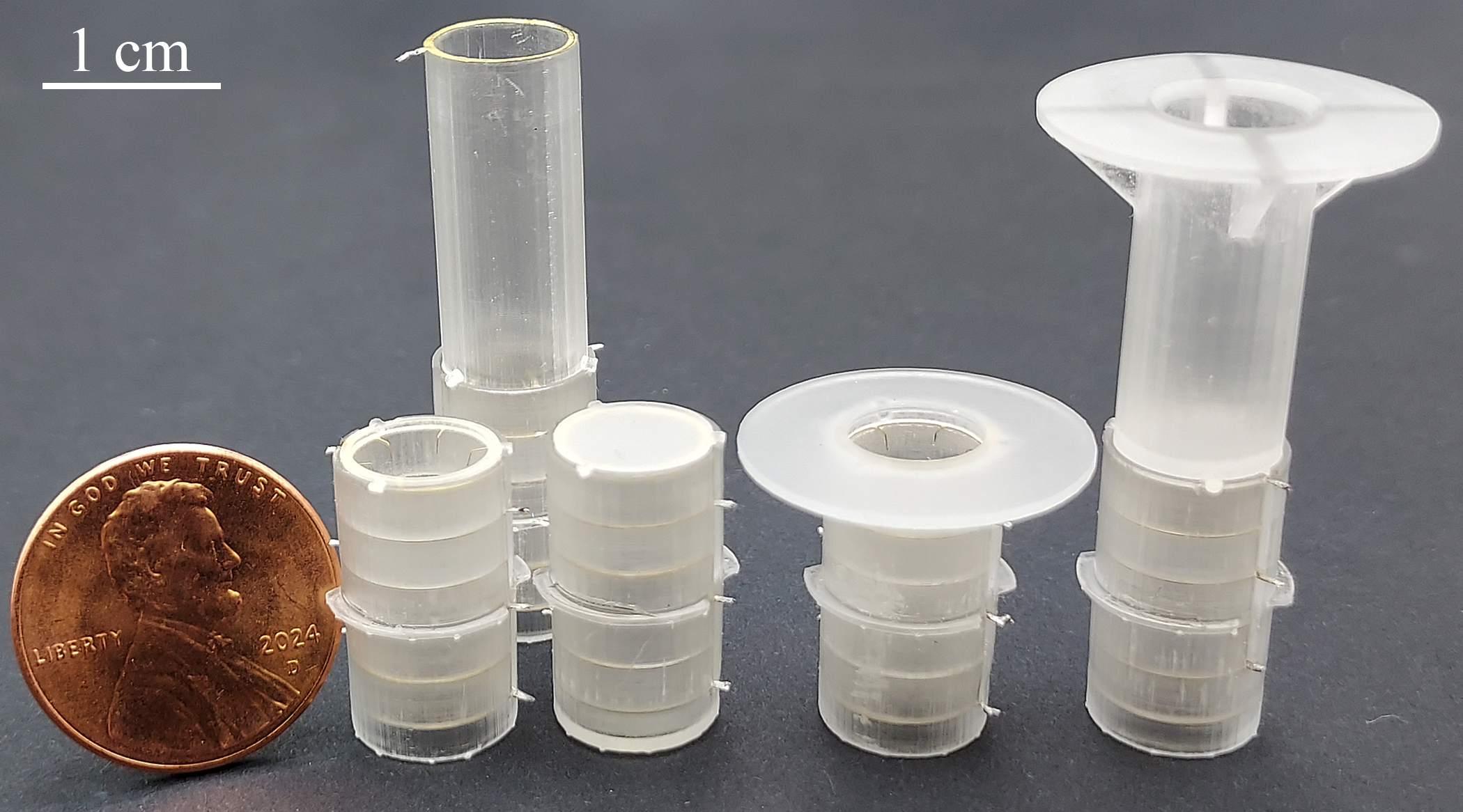}
    \vspace{-6mm}
    \caption{Thrusters used in experiments. Devices from left to right: control ``standard'' device, extended intake duct with grounding ring for isolating electrostatic adhesion, fully plugged device for isolating electrostatic effects, flanged device for increasing electrostatic and fluid mechanic effects, and flanged device with extended intake duct for isolating fluid mechanic effects.}
    \label{fig:1_teaser}
    \vspace{-4mm}
\end{figure}

\begin{figure}
    \centering
    \includegraphics[width=0.9\columnwidth]{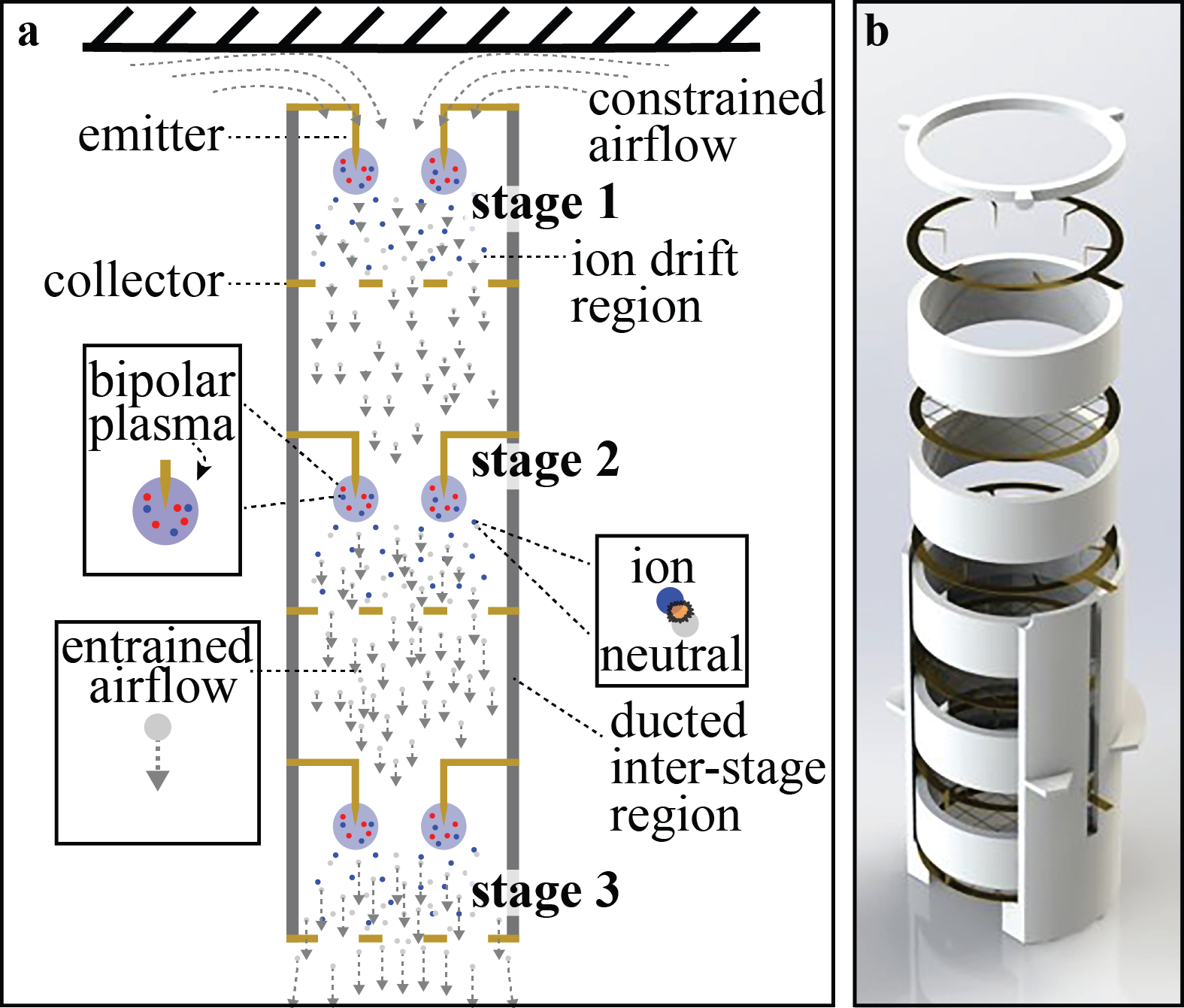}
    \vspace{-2mm}
    \caption{\textbf{(a)} Schematic view of a three-stage electroaerodynamic ducted actuator near a ceiling plane. A high potential applied between the ``emitter'' and ``collector'' electrodes ignites a local plasma from which ions are ejected. These ions drift in the electric field towards the collector, and frequent momentum-transferring collisions with neutral air molecules along the way result in an entrained air jet. Successive stages increase the velocity of the air as it passes through the duct. \textbf{(b)} Render of the three-stage ducted devices used in this work, where UV-laser micromachined brass electrodes are integrated with SLA-printed ducts in an adhesiveless process. Design, fabrication, and assembly of these devices is drawn from prior work~\cite{nelson2024high}.}
    \label{fig:2_schematic}
    \vspace{-2mm}
\end{figure}

EAD propulsion is generated by momentum-transferring collisions between ions, accelerated in an electric field, and air molecules (Fig.~\ref{fig:2_schematic}a).  Centimeter-scale robots using EAD actuators (Fig.~\ref{fig:1_teaser}) have been shown to fly, roll, hop, or perform a combination of these movements~\cite{drew2018toward,zhang2022centimeter}. MAV-scale vehicles which can carry useful payloads or move with onboard power, however, remain elusive due to limited thrust efficiency. A potential solution is to leverage the significant increase in apparent thrust which is achievable through operation in close proximity to a surface. Our recent work showed, for the first time, that a large increase in thrust and thrust efficiency is possible for ducted multi-stage EAD actuators operated close to a ground plane about the outlet~\cite{nations2024empirical}. The presence and magnitude of a similar effect when operated close to a ``ceiling'' plane, until this work, was uncertain; it has never been shown before for EAD.

While endeavoring to perform the first empirical investigation into ceiling effects for EAD thrusters, we discovered a surprisingly high magnitude electrostatic attractive (ESA) force near the ceiling plane. Based on the design of our devices, it suggests that the EAD thruster's dielectric structure may be experiencing electrical polarization and surface charge build-up from the corona plasma~\cite{monkman1997analysis}. This works in tandem with the electrostatically-induced polarization between the ceiling plane and high-voltage emitter electrode (and edge of the corona plasma sheath), more analogous to traditional electroadhesive devices, to create a significant attractive force which could not be predicted from just the exposed surface area of the high voltage electrode. This force is shown to combine with aerodynamic benefits, leading to a higher magnitude of ceiling proximity effects as compared to small scale rotorcraft or flapping wing MAVs~\cite{yamamoto2007wall}.

Predicting the benefits of the combined aerodynamic and electroadhesive forces is challenging. The expressions typically used for modeling EAD thrust (Eq.~\ref{eq:idmu}) and thrust efficiency (Eq.~\ref{eq:eta}), where $\mu$ represents ion mobility, $E$ is the average drift field magnitude, and $v$ is the free-stream velocity, are decoupled from aerodynamics and rely on a simplified one-dimensional theory for space charge-limited discharges~\cite{pekker2011model, masuyama2013performance}. Electroadhesion itself is also difficult to predict analytically or numerically, with various hyperparameters (e.g., related to both material and electrode geometry) typically used to fit empirical values to equations~\cite{chen2013modeling}.

\vspace{-1em}
\begin{center}
\begin{tabular}{p{4cm}p{3cm}}
  \begin{equation}
  F = \frac{Id}{\mu} = \frac{9}{8} \epsilon_0 A E^2
  \label{eq:idmu}
  \end{equation}
  &
  \begin{equation}
  \eta = \frac{1}{\mu E + v}
  \label{eq:eta}
  \end{equation} 
\end{tabular}
\vspace{-1em}
\end{center}

The contribution of this work is \textbf{the first-ever empirical investigation of ceiling proximity effects for electroaerodynamic thrusters}. We examine the impact of factors including the distance of the thruster inlet from the ceiling plane, different ceiling materials, and variable inlet flange widths (Fig.~\ref{fig:3_4_devices_setup}b). We also devise a method for isolating and measuring the electroaerodynamic (i.e., fluid mechanic) and electrostatic attractive (i.e., electroadhesive) forces separately (Fig.~\ref{fig:6_glass_conductive}). Our results demonstrate that EAD ceiling proximity effects can result in a large gain in thrust and thrust efficiency, up to $600\%$ in certain configurations, and that the addition of an intake flange can result in a net increase in payload capacity despite its added mass for certain sizes of flange. The magnitude of this effect indicates that it should not be overlooked in future studies of EAD-propelled MAVs intended for indoor operation, and opens the door to entirely new designs for high-efficiency vehicles capable of utilizing it for efficient locomotion near a ceiling plane and for perching.

\section{Background and Related Work}

\subsection{Physical Basis of Ceiling Proximity Effects}
Aerodynamic proximity effects refer to the fluid dynamic forces and flow patterns that occur when a fixed wing, rotor, or flapping wing device operates in close proximity to a boundary, such as the ground, a ceiling, or a wall. Ceiling proximity effects occur when operating near a plane close to the actuator inlet---not necessary an overhead surface. There has been limited research in this area, as the primary focus of past investigations into proximity effects has been on studying helicopters or large fixed-wing aircraft at high Reynolds numbers outdoors, which pose safety concerns when operated near surfaces and don't typically come near ceilings~\cite{hsiao2019ceiling}. Only in the past several years, with the increasing use of MAVs indoors and in confined spaces, has the study of ceiling proximity effects attracted interest~\cite{jimenez2019contact}. 

When flying near an overhead surface, known as in-ceiling-effect (ICE), the fluid flow is obstructed by the surface, leading to a reduction in the downwash force on wings and rotors which results in increased thrust~\cite{carter2021influence}. Operation ICE also induces a greater pressure gradient across the wings or rotors compared to the pressure gradient that exists in out-of-ceiling-effect (OCE) conditions. This pressure gradient creates a low-pressure zone between the actuator and the ceiling, with a high-pressure zone below, causing the vehicle to be pulled towards the ceiling. If controlled effectively, it can be harnessed to improve flight efficiency~\cite{kocer2019aerial}.

\subsection{Ceiling Effects for Small Rotorcraft}
Ceiling proximity effects are typically characterized as a function of the dimensionless parameter distance-to-rotor radius, $z/r$, where $z$ represents the distance between the ceiling plane and the thruster’s inlet, and $r$ represents the actuator radius (Fig.~\ref{fig:3_4_devices_setup}b). A major benefit noted by prior studies of the ceiling effect in rotorcraft is increased efficiency. For example, Hsaiao and Chirarattananon predicted that by leveraging measured ceiling effects, required power for hover could be reduced by a factor between two and three~\cite{hsiao2018ceiling}. Kocer et al. and Gao et al. demonstrated reduced power consumption ICE~\cite{kocer2019uav},~\cite{gao2019exploiting} of around 15\%, resulting in a corresponding increase in flight time. Blade element theory combined with empirically-derived fit parameters yields reasonably accurate models for the ceiling effect in the regimes where the empirical values hold~\cite{du2024modelling,conyers2018empirical}. This approach has been extended with numerical simulation to improve accuracy~\cite{jinjing2023aerodynamic,liu2024analysis}. Incorporating ceiling effect predictions into the motion planning loop has been shown to improve flight safety and accuracy by various studies~\cite{kocer2019aerial,jimenez2019contact,gao2019exploiting}. 

Our work has the clear distinction of using electroaerodynamic (EAD) thrusters rather than rotors. As the thrusters are static and utilize ion collisions with neutral air molecules to generate thrust, blade element theory is not applicable, and therefore neither are the many studies combining that analytical approach with empirically-derived fit parameters for reduced scale and Reynolds number. Our work also explores the smallest diameter propulsor among the reviewed studies. This implies that, for our devices, a low $z/r$ ratio translates to a comparatively smaller physical distance from the ceiling plane.  This study is the first step towards developing and validating a reliable predictive model for fully describing the aerodynamics of EAD thrusters ICE.

\subsection{Physical Basis of Electrostatic Adhesion}
Electrostatic adhesion, also known as electroadhesion, is the electrostatic attractive normal force generated between two materials, typically a dielectric in close proximity to conductive electrodes with a high surface charge density, based on polarization of the dielectric from the high magnitude electric field~\cite{krape1968applications}. Various types of electrodes have been reported for electroadhesion applications, with interdigitated electrodes embedded in a dielectric to prevent discharge during contact being the most common type~\cite{chen2013modeling,bamber2017visualization}. As the electric potential between the electrodes increases, a higher electric field is generated, leading to a higher surface charge density and increased attractive force~\cite{tellez2011characterization}. Electroadhesion is strongly dependent on environmental parameters and is difficult to predict analytically or numerically~\cite{chen2013modeling}.

To our knowledge, this is the first time an electroadhesive force has been reported from a corona discharge-based device. The goal of our devices is to generate a stable atmospheric plasma around the ``emitter'' electrode. To do so, a high magnitude electric field must be formed in the vicinity of the emitter. Based on the electrode geometric asymmetry, the field shape leads to a high magnitude fringe above the emitter (i.e., extending out of the inlet of the device). We posit that this field is what is inducing polarization of the ceiling plane and subsequent electroadhesion. Based on the results of our attempts to isolate this force, we believe there is also a component of force arising from charging of the dielectric surface of the duct by ions ejected from the plasma, which is then attracted to the polarized ceiling plane. 

\subsection{Electrostatic Adhesion for Robotics}
Electrostatic adhesion has been used in robotics for decades~\cite{guo2019electroadhesion}. It is an attractive mechanism for manipulation, for example, due to its ability to handle fragile components and a wide variety of material surfaces in diverse environments, including in a vacuum~\cite{bamber2017visualization}. Additional advantages include being low-noise, lightweight, flexible, efficient, and offering switchable adhesion for catch and release~\cite{prahlad2008electroadhesive}. Many versatile robotic grippers based on the principle of electroadhesion have been demonstrated in recent literature~\cite{shintake2016versatile,chen2022variable,hwang2021electroadhesion}. An emerging area of interest for electroadhesion is effective perching and climbing of small-scale robots, which benefit from increased surface area-to-volume ratios and have unique design constraints necessitating simple, lightweight mechanisms. Climbing robots have been shown that can adhere to various material surfaces~\cite{prahlad2008electroadhesive,de2018inverted}, and it has been used as a perching mechanisms for both flapping wing MAVs~\cite{graule2016perching} and rotorcraft~\cite{park2020lightweight} to extend battery life. 

\section{Methods}
\label{sec:methods}
\vspace{-2mm}
\subsection{Thrusters}
The thrusters used in this study are similar to those recently presented in \cite{nelson2024high} and \cite{nations2024empirical}, with modifications made to assist in isolation of EAD and ESA forces. A thruster comprises three ducted, stacked stages (Fig. \ref{fig:2_schematic}), each with a height of 3 mm. The device has an overall height of 17.6 mm (not considering extended intake duct devices, which are discussed later in this section), an inner diameter of 8 mm, and an outer diameter of 10 mm. Duct structures are stereolithographically (SLA) printed, the brass electrodes are UV-laser micromachined, and all components are secured in place using an adhesiveless press-fit cap. Further details are omitted here for brevity, and the reader is directed to~\cite{nelson2024high} for a full discussion of these devices. 

\begin{figure}
    \centering
    \includegraphics[width=\columnwidth]{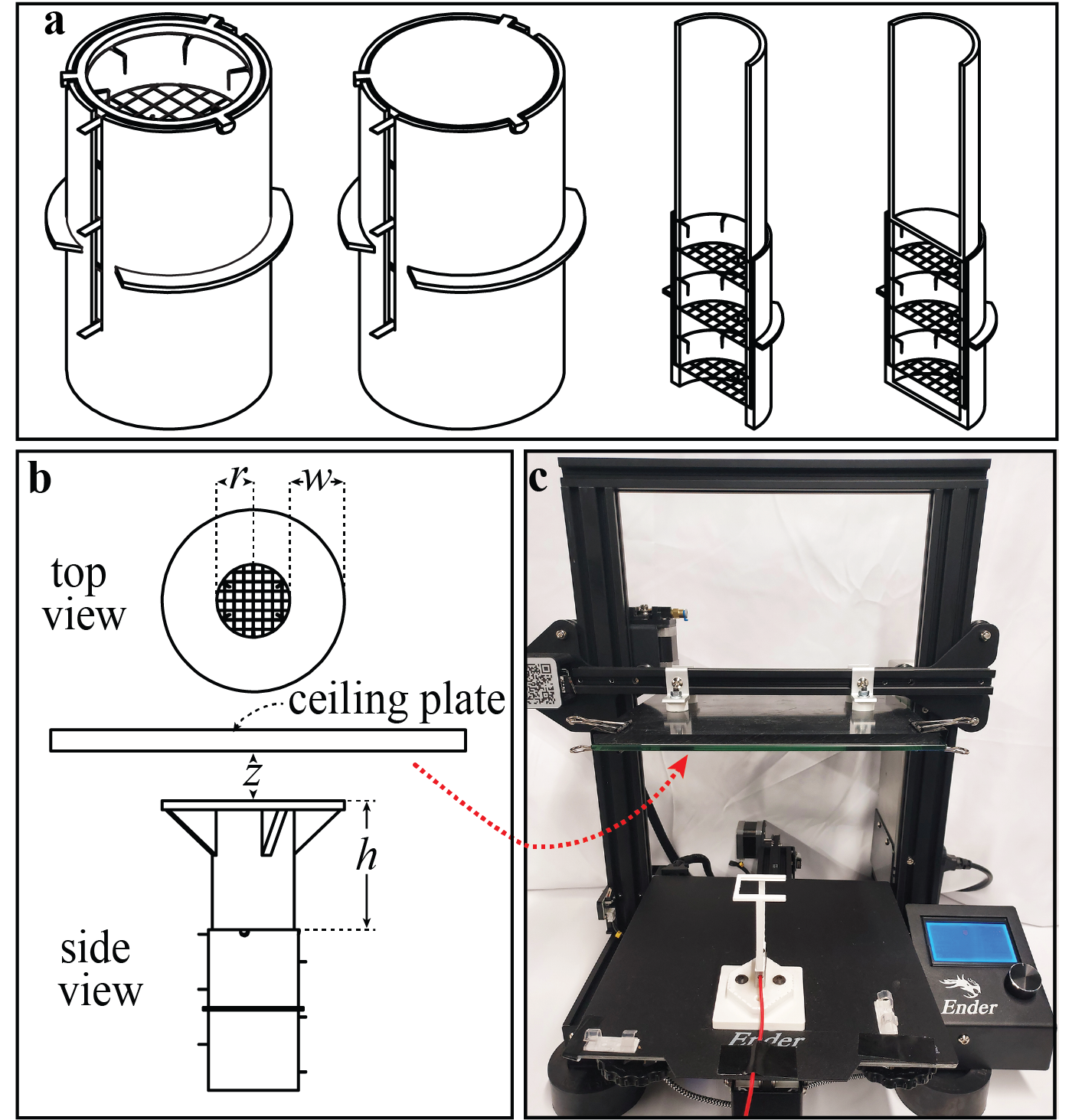}
    \vspace{-6mm}
    \caption{\textbf{(a)} From left to right: A standard device, plugged device, and cross sections of open and plugged devices with extended intakes. \textbf{(b)} Geometric parameters of interest, including the flange width $w$, thruster radius $r$, thruster nozzle height from the top emitter $h$ for electrostatic isolation, and the distance from the thruster inlet to the ceiling plate $z$. \textbf{(c)} The automated test setup constructed from a modified 3D printer synchronizes data acquisition with precise movement of the ceiling plate.}
    \label{fig:3_4_devices_setup}
    \vspace{-4mm}
\end{figure}

Motivated by experimental results, several modifications have been made to the thruster design to isolate EAD and ESA forces. A ``standard'' device (Fig. \ref{fig:3_4_devices_setup}a) is identical to those used in \cite{nations2024empirical}, which made slight structural modifications to their original presentation by Nelson and Drew \cite{nelson2024high}. The first modification is to ``plug'' a device by closing the inner diameter of the thruster exhaust and the inner diameter of the cap, which is the mechanism above the uppermost stage emitters that secures all the device's components in place. The second modification is using an extended intake duct above the first emitter stage, whose height is denoted as $h$ in Fig.~\ref{fig:3_4_devices_setup}b. A device with an extended intake may be open or plugged. Notably, the ``plug'' in the extended cap is toward the emitter side of the extension, so that the surface area facing the ceiling plate remains the same as an open extended intake device. The process through which changes are used to isolate EAD and ESA forces is discussed in Section \ref{sec:results}.

\subsection{Experimental Setup}

As in \cite{nations2024empirical}, the experimental testbed used in this study is constructed from a modified Ender 3 Pro 3D printer (Fig.~\ref{fig:3_4_devices_setup}c). The X-axis stepper and extruder are replaced by a 230 mm by 230 mm Delrin acetal plastic plate secured to the aluminum frame which can be raised or lowered to the desired height above the thruster ($z$ in Fig.~\ref{fig:3_4_devices_setup}b) using G-code sent from a Python script. We tested two additional ceiling materials, glass and conductive fabric tape, where the glass plate is a borosilicate glass 3D printer bed, and the conductive tape surface is the same glass plate with one side covered uniformly by two overlapping layers of Faraday cloth tape, by mounting additional plates under the Delrin. The device under test is secured in a mount (Fig.~\ref{fig:3_4_devices_setup}c) designed for unobstructed airflow, printed using Formlabs Rigid 10K resin. The mount has a height of 53 mm, placing the exhaust approximately 40 mm from the FUTEK LSB200 S-Beam load cell to which the mount is attached. The device is oriented so the intake points toward the ceiling plate. 

Voltage is applied to devices using a Spellman SL10P30 programmable high-voltage power supply with a separate Python-controlled variable power supply connected to its remote inputs. Its output monitor gives access to real-time voltage readings, which are recorded along with measurements across a shunt resistor on the collector side of the EAD device with a Rigol DS1054Z oscilloscope throughout each experimental trial (a voltage sweep up to $\approx \!\!$ 3 kV) to generate paired voltage and current data points.


\section{Results and Discussion}
\label{sec:results}
 Experiments are performed with two devices per condition and three trials per device. Plotted data is the mean of each device's trial mean with standard error of the mean (SEM) error bars. The error of the numerator is used for ratios. A glass ceiling plane is used unless otherwise stated.

\subsection{Isolating Constituent Forces}


We identified a significant electroadhesive force when a thruster is ICE. Isolating the EAD and ESA forces is challenging because changes made to eliminate one often affect the other (e.g., blocking airflow with a plug increases surface area for charging). The method we arrived at relies on the numerous thruster configurations discussed in Section~\ref{sec:methods} and is in part algebraic, assuming that the total force produced by the thruster is the electrostatic adhesive force ($F_{\text{ESA}}$) and electroaerodynamic force ($F_{\text{EAD}}$), with aerodynamic drag included in the EAD term: $F_{\text{tot}} = F_{\text{ESA}} + F_{\text{EAD}}$. 

\begin{figure}
    \centering
    \includegraphics[width=\columnwidth]{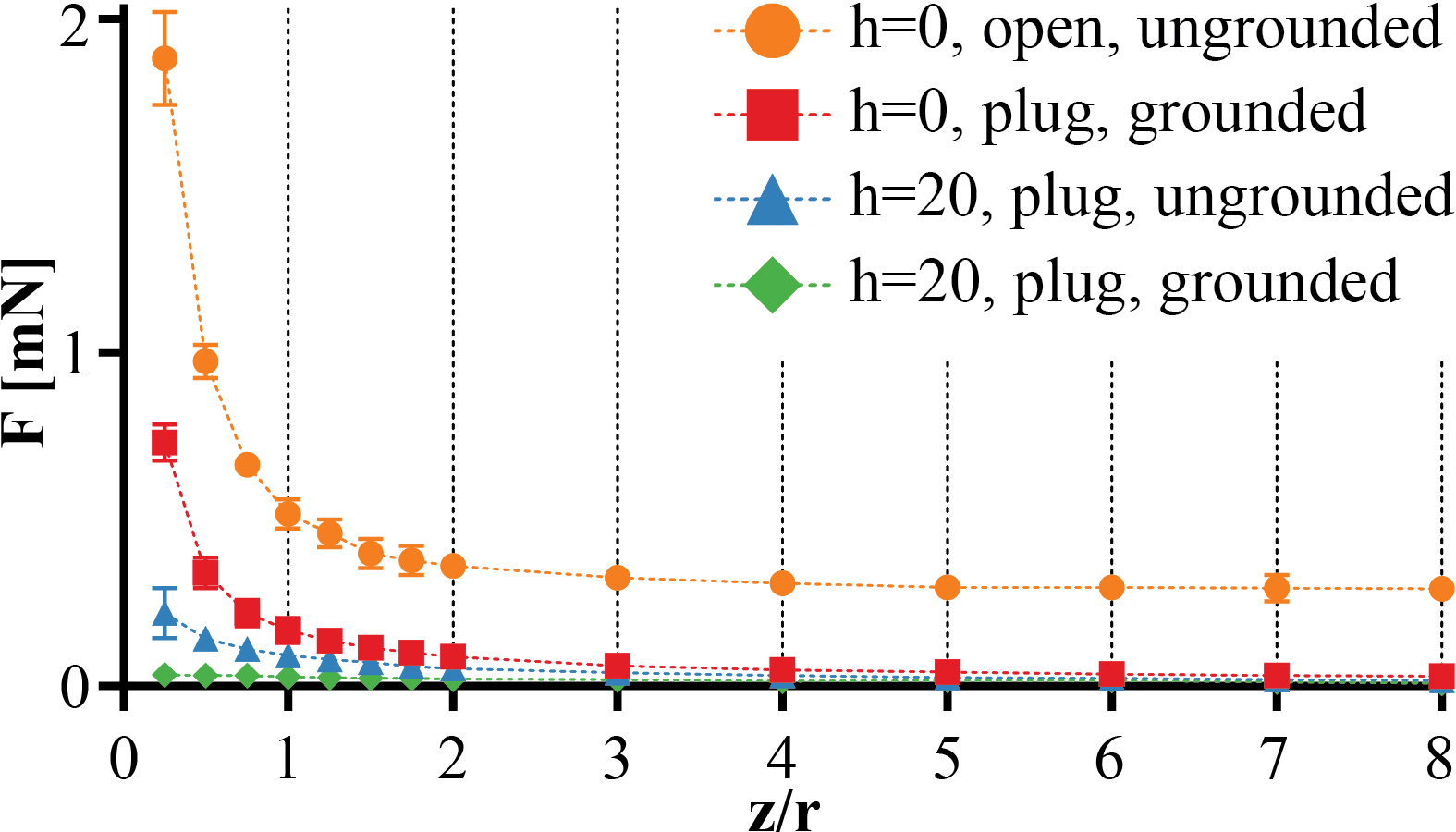}
    \vspace{-6mm}
    \caption{Effect of device configuration on produced force versus distance from the ceiling plane. The $h=0$, open, ungrounded configuration represents a ``standard'' device, while we show that a device with an extended intake duct, an airflow plug above the first emitter stage, and a grounding ring on the intake annulus ($h=20$, plug, grounded), results in zero force at any distance. Collected with conductive tape on glass as the ceiling plane.}
    \label{fig:5_EAD_isolation}
    \vspace{-2mm}
\end{figure}

To isolate EAD force, we use plugged and open variations of an extended intake device, with the plug located directly above the first emitter stage in order to not increase the surface area in close proximity to the ceiling plate (see Fig.~\ref{fig:3_4_devices_setup}a). The height of the extended intake is 20 mm ($z/r$ = 5), which, for a standard device, is OCE (Fig. \ref{fig:5_EAD_isolation}: $h=0$, open, ungrounded). This distance mitigates any influence from the device's electric field and has a negligible impact on surface drag (less than 1$\%$ measured difference OCE). That this electric field is present above the first stage emitters and a significant contributor to the ESA force is shown in experiments with a plugged device \textit{without} an extended intake duct, with a copper lead connecting the outer surface of the cap to a grounded ceiling plate (Fig. \ref{fig:5_EAD_isolation}: $h=0$, plug, grounded) which prevents  charge accumulation on the thruster exterior.

Experiments with the extended intake duct device (Fig. \ref{fig:5_EAD_isolation}: $h=20$, plug, ungrounded) show a dramatic reduction in ESA force, with the remainder hypothesized as being due to surface charge accumulation on the duct annulus by ions ejected from the corona plasma. This is supported by experiments with a plugged extended intake device whose intake lip has an approximately \SI{25}{\micro\meter} thick brass ring secured to it using Kapton tape, which is further connected to the conductive ceiling plate via a copper lead. This acts as a ``ground ring'', and eliminates all remaining ESA attraction to the ceiling plane (Fig. \ref{fig:5_EAD_isolation}: $h=20$, plug, grounded).

\begin{figure}[t]
    \centering
    \includegraphics[width=\columnwidth]{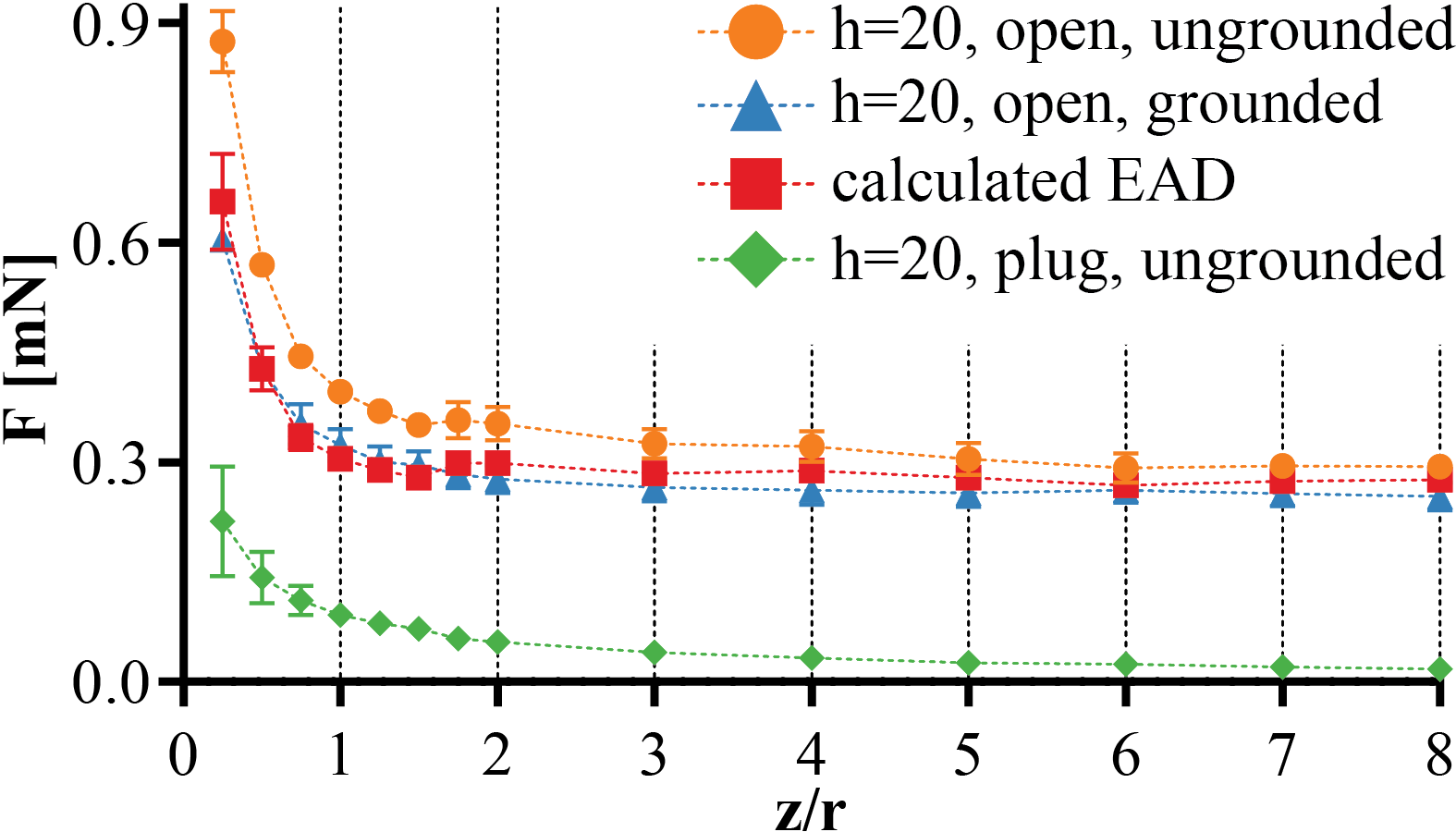}
    \vspace{-6mm}
    \caption{Demonstration of isolating electroaerodynamic force from electrostatic attractive force. The $h=20$ plugged device data is subtracted from the $h=20$ open device data to yield a calculated pure-EAD force. This is confirmed by agreement between this calculated value and the $h=20$, open, grounded device data. Collected with a conductive tape on glass ceiling.}
    \label{fig:6_glass_conductive}
    \vspace{-2mm}
\end{figure}

\begin{figure}
    \centering
    \includegraphics[width=\columnwidth]{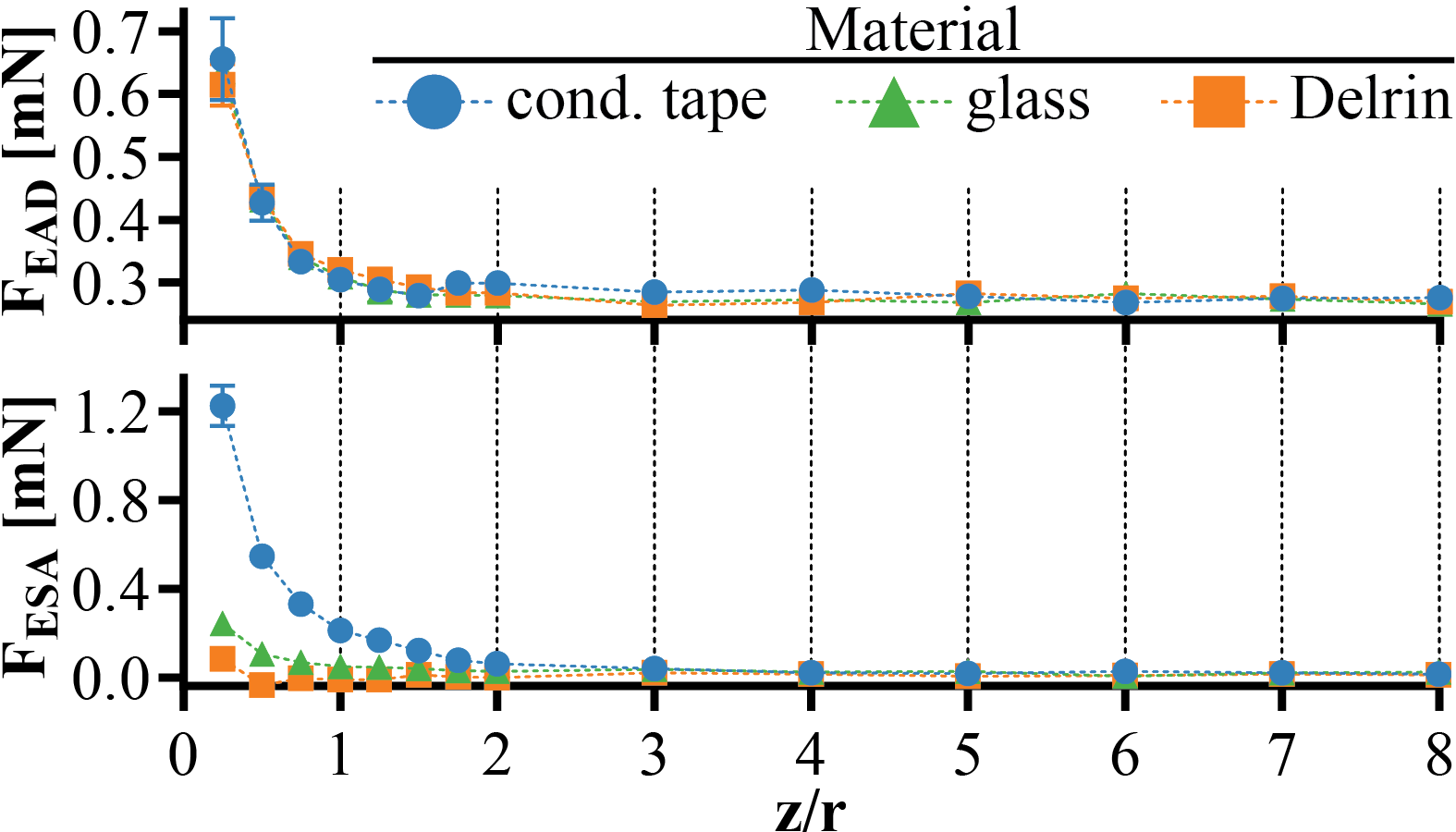}
    \vspace{-6mm}
    \caption{\textbf{Top}: Isolated electroaerodynamic force versus distance from the ceiling plate for different materials. \textbf{Bottom}: Isolated electrostatic attractive force versus distance from the ceiling plate for the same materials.}
    \label{fig:7_8_material}
    \vspace{-3mm}
\end{figure}

The EAD force of a ``standard'' thruster is then found by subtracting the force of a plugged extended intake device from that of an open extended intake device, removing the electrostatic attractive component. Similarly, $F_{\text{EAD}}$ may be recorded experimentally with an open extended intake device whose intake lip is grounded using the same brass ring and conductive ceiling discussed previously (Fig. \ref{fig:6_glass_conductive}: $h=20$, open, grounded). Figure \ref{fig:6_glass_conductive} shows that both methods produce nearly identical data, which supports the algebraic method of arriving at this measurement. This method is useful in that it is material agnostic; we show in Figure \ref{fig:7_8_material} that between conductive fabric, glass, and Delrin, the EAD force remains similar, with the minor difference between materials explainable by variations in surface roughness. The calculated ESA force for different ceiling materials, found by  subtracting the isolated EAD force from the total force, agrees with  trends predicted by literature, where the conductive fabric surface has the highest magnitude, followed by glass, and lastly the Delrin~\cite{guo2015investigation,guo2017experimental} (Fig. \ref{fig:7_8_material}).

\subsection{Thrust and Thrust Efficiency of a Standard Device}

\begin{figure}
    \centering
    \includegraphics[width=\columnwidth]{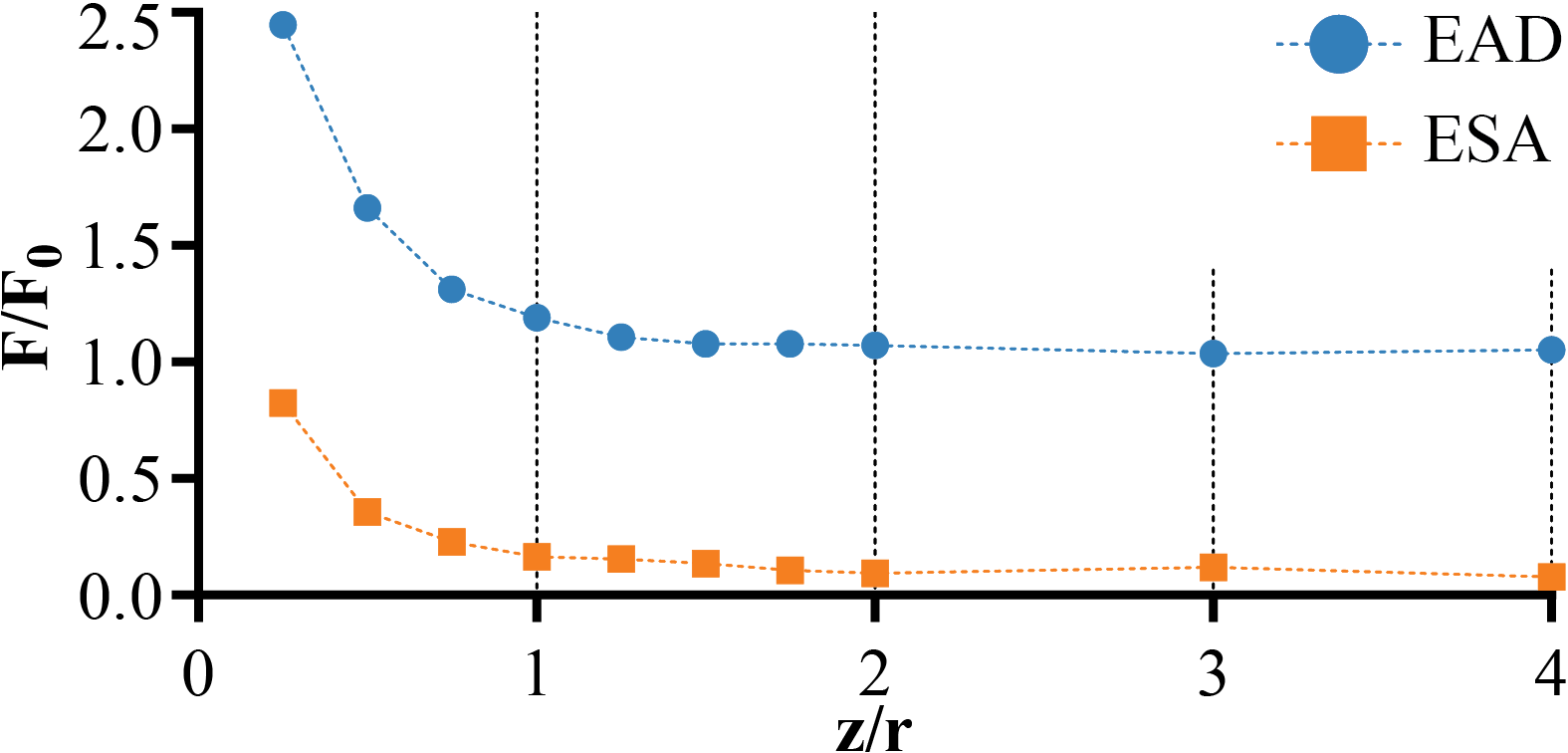}
    \vspace{-6mm}
    \caption{The electroaerodynamic (EAD) and electrostatic attractive (ESA) components of the effective thrust relative to the OCE thrust as a function of distance from the ceiling plane. $F_0$ is of a standard device OCE.}
    \label{fig:9_glass_relative_force}
    \vspace{-2mm}
\end{figure}

Close proximity of a standard thruster to a glass ceiling plane results in an effective thrust (i.e., combined ESA and EAD forces) of up to 0.88 mN, an approximately 200\% increase over OCE thrust (0.29 mN). Using the methods outlined in the previous section, approximately 80$\%$ of this increase can be attributed to the increased electroaerodynamic force, with the remaining 20$\%$ due to electrostatic attraction (Fig. \ref{fig:9_glass_relative_force}). As is expected, as the thruster approaches an OCE $z/r$ of 4, the ESA force goes to zero while the relative EAD force approaches unity. 

\begin{figure}
    \centering
    \includegraphics[width=\columnwidth]{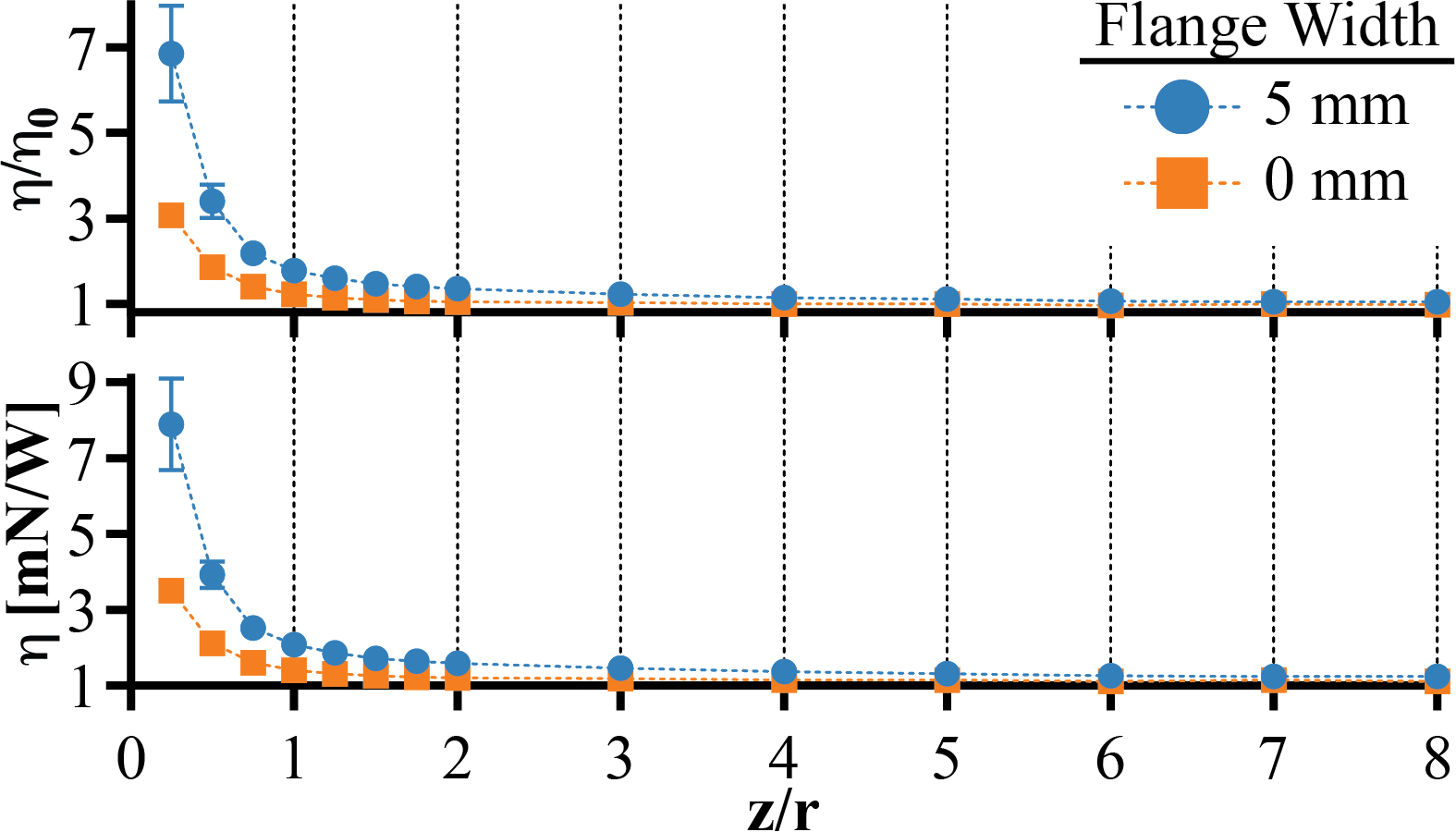}
    \vspace{-6mm}
    \caption{The relative thrust efficiency and measured thrust efficiency versus $z/r$ for a standard device with flange widths of 5 mm and 0 mm.}
    \label{fig:11_efficiency_rel_abs}
    \vspace{-3mm}
\end{figure}

\begin{figure}
    \centering
    \includegraphics[width=\columnwidth]{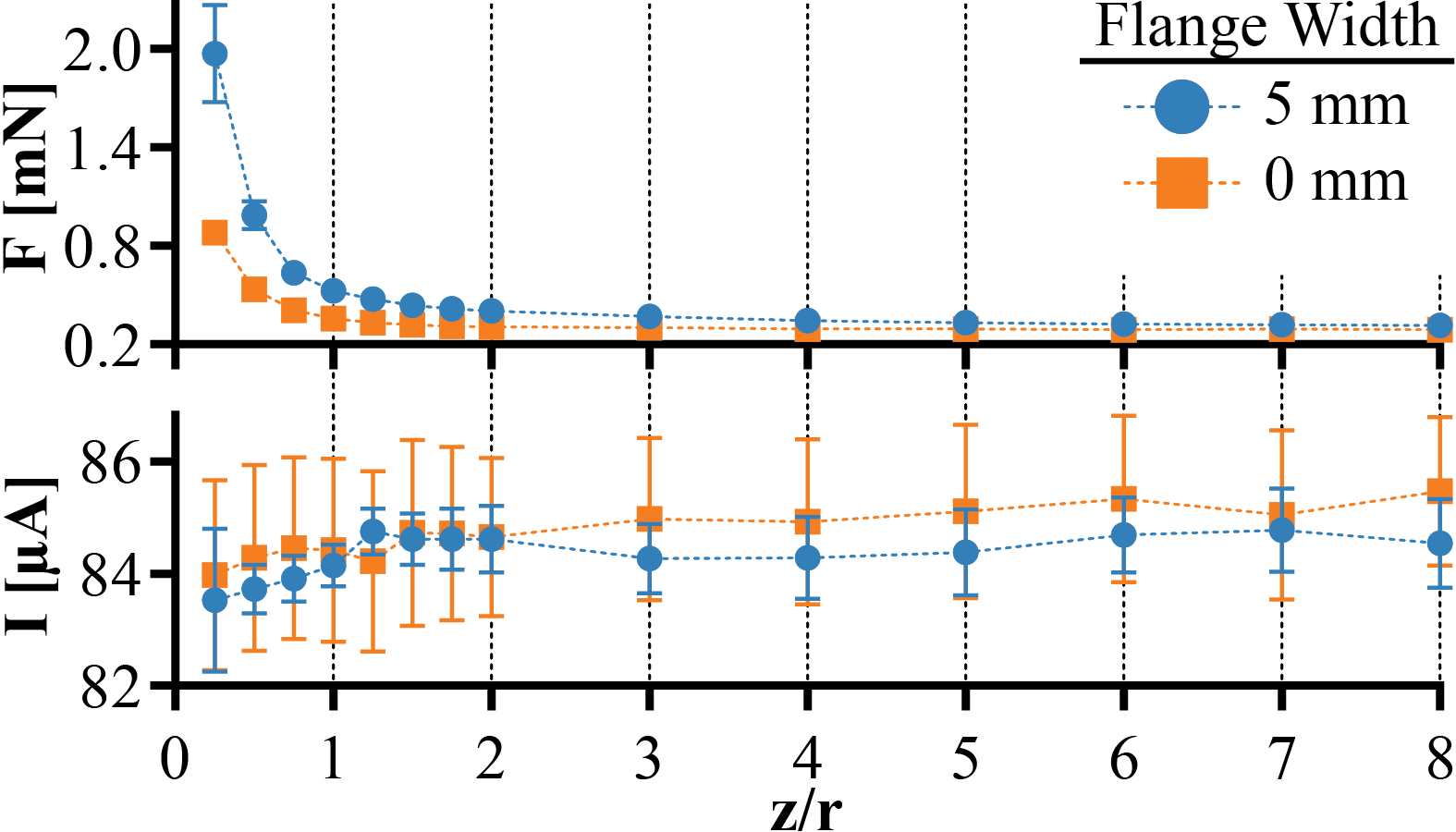}
    \vspace{-6mm}
    \caption{The measured force and current versus $z/r$ for a standard device with flange widths of 5 mm and 0 mm.}
    \label{fig:10_current_force_smoosh}
    \vspace{-2mm}
\end{figure}

The thrust efficiency of the standard thruster with a glass ceiling plane shows a similar increase in efficiency of approximately 200$\%$ at a distance of 1 mm to the plane (Fig.~\ref{fig:11_efficiency_rel_abs}, 0~mm flange width). Interestingly, we see a slight \textit{decrease} in current at low $z$ ($\leq$4 mm), which serves to benefit efficiency (Fig. \ref{fig:10_current_force_smoosh}, bottom). This stands in contrast to previous findings with EAD thrusters in ground effect, which shows current to slightly \textit{increase} close to the ground plane~\cite{nations2024empirical}. One hypothesis is that the stagnation point introduced between the inlet and ceiling serves to decrease the total effective air velocity ICE relative to OCE, in contrast with the ground effect where the stagnation point is below the device and the increase was attributed to a larger convective current term from the increased fluid velocity. It is challenging to confirm or test our hypotheses about this effect without something like particle image velocimetry (PIV) for direct visualization of inlet and outlet velocities, which remains as future work.

\subsection{Intake Flanges}
\label{ssec:intake_flanges}

Motivated by our past work which showed that the addition of circular exhaust flanges altered ground proximity effects in EAD thrusters \cite{nations2024empirical}, we investigated incorporation of circular flanges on the thruster inlets. Our results show that intake flanges have significant effects on both the EAD and ESA force components ICE. A device with a 5 mm intake flange was found to increase efficiency by up to 600\% ICE, achieving a maximum recorded efficiency of 8 mN/W (Fig. \ref{fig:11_efficiency_rel_abs}). The increase in thrust is similar, with a maximum recorded effective thrust of 2 mN ICE, compared to 0.3 mN OCE (Fig. \ref{fig:10_current_force_smoosh}). These are similar relative increases to the values measured for the ground effect~\cite{nations2024empirical}. 

\begin{figure}
    \centering
    \includegraphics[width=\columnwidth]{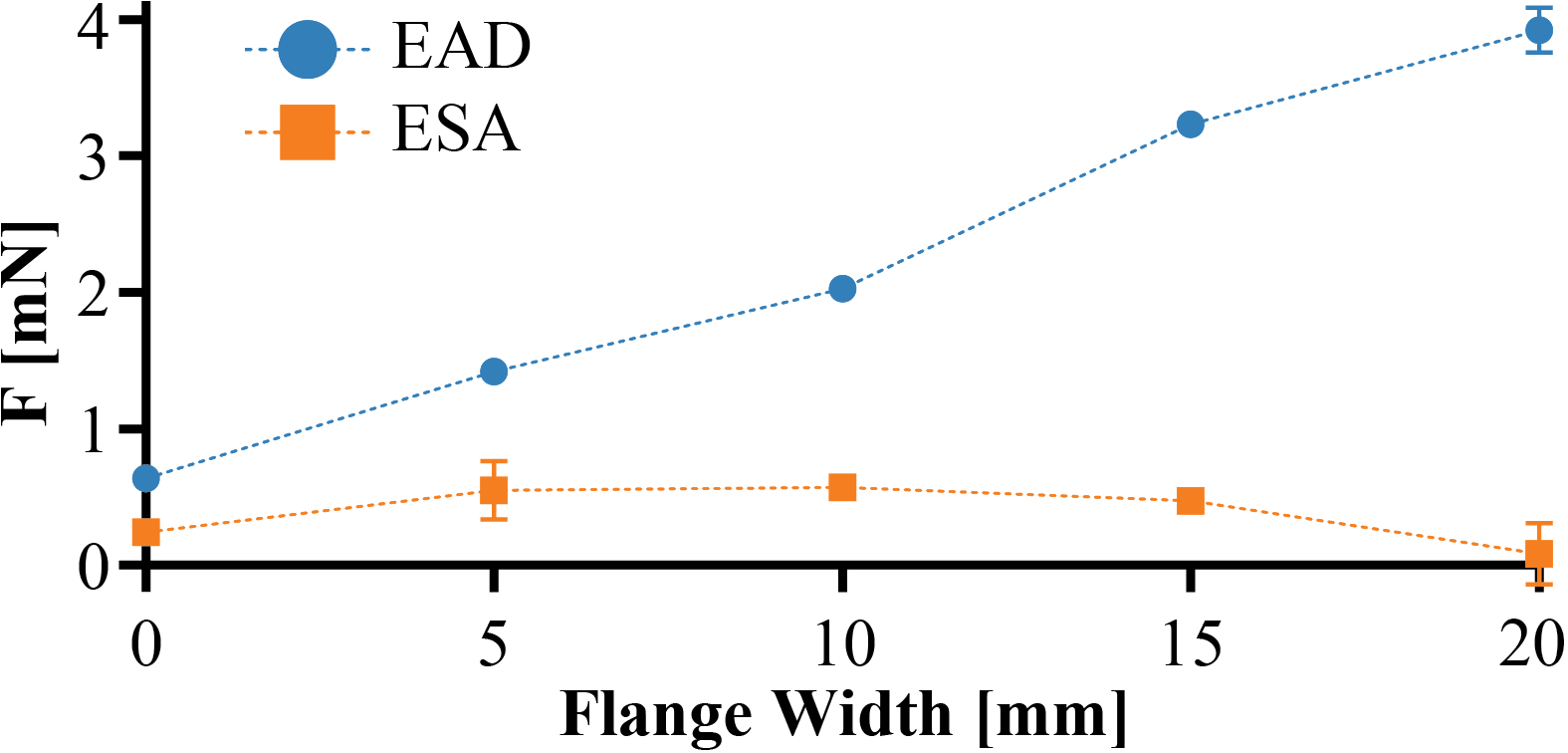}
    \vspace{-6mm}
    \caption{Measured electroaerodynamic (EAD) and electrostatic attractive (ESA) force components versus flange width at $z$ = 1 mm.}
    \label{fig:12_flange_EAD_ESA}
    \vspace{-3mm}
\end{figure}

EAD force was shown to monotonically increase as a function of flange width (5, 10, 15, and 20 mm) at a distance of 1 mm from the ceiling (the point of maximum recorded effective thrust), while ESA force was shown to increase until a flange width of 10 mm and then decrease (Fig. \ref{fig:12_flange_EAD_ESA}). The increase in EAD force with increasing flange width makes sense, as it increases the surface area. We assume that the fluid velocity is increased close to the ceiling plate relative to an OCE device, resulting in a pressure differential and an apparent lift across the flange. One hypothesis for the increase in electrostatic attractive force is that the increased flange width presents a larger surface area to be charged by ions emitted from the corona plasma. At some point, the surface area is large enough where this surface charge can disperse over such an area that their ability to polarize the glass ceiling plane decreases or where the ceiling plane is no longer polarized by the electric field near the charges.

\begin{figure}
    \centering
    \includegraphics[width=\columnwidth]{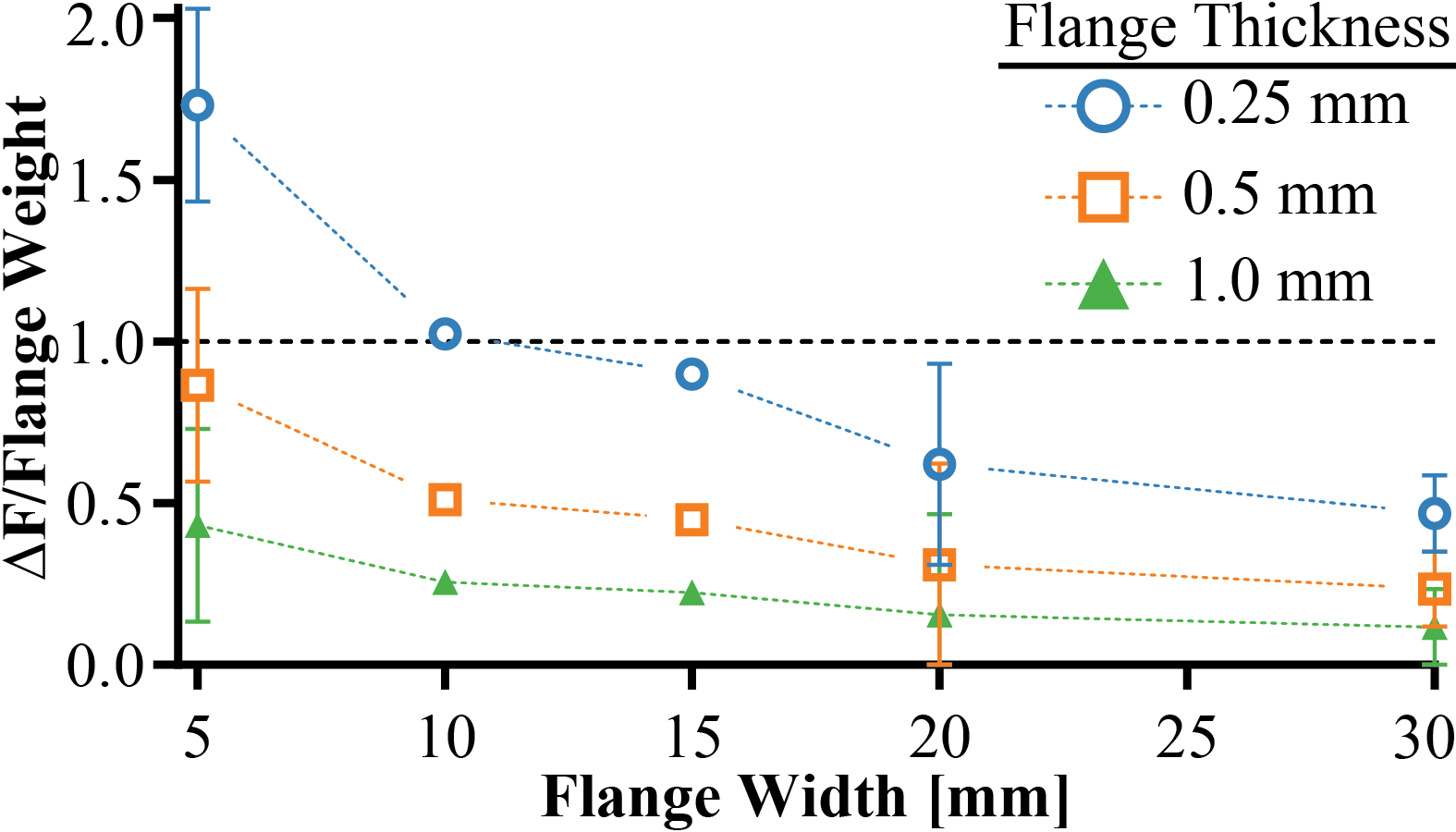}
    \vspace{-6mm}
    \caption{Measured (filled marker) and potential (open marker, calculated) marginal increase in force versus flange width for different thicknesses.}
    \label{fig:13_del_fwr_v_flange_hollow}
    \vspace{-4mm}
\end{figure}

We also considered the change in force relative to the weight of the flange as a function of the flange material thickness in Fig.~\ref{fig:13_del_fwr_v_flange_hollow}. This value monotonically decreases with increasing flange width, and we see that for our printed 1 mm thick flanges, it is never beneficial from a thrust-to-weight standpoint to add a flange. Recalculating the flange weight using different potential material thicknesses, however, we see that this is a beneficial effect for thicknesses of 0.25 mm or lower (e.g., achievable by Kapton or PEI). 
Notably, while a flange increases the total thrust and thrust efficiency, it decreases the thrust density; a device with no flange achieves a thrust density of approximately 17.5 $N/m^2$ ICE, while one with a 5 mm flange achieves only about 8 $N/m^2$. 

\section{Limitations and Future Work}
There were several limitations to this study. We did not consider the Reynolds number dependence of the ceiling proximity effect; in prior work we found that the magnitude of ground effect was independent of Reynolds number over the limited range achievable by these thrusters, but at higher achievable Reynolds numbers this is likely to be significant. While we explored the electrostatic attractive force component of the ceiling effect, we did not directly measure the electroadhesive pressure as done traditionally for gripping and perching applications (i.e., using pull-off or peel force); because the ESA force of our devices is mediated by a combination of electric field and surface charging, it is unclear how contact would affect it. Finally, we did not attempt to reconcile existing models for ceiling proximity effects which take advantage of blade element theory with our data due to fundamentally different underlying physics; data-driven modeling using our current data would be premature, and physics-based modeling is outside the scope of this paper. In the future, there are several other device parameters which should also be investigated, including multi-thruster interactions, thruster radius, forward flight speed (or ambient flow velocity), and angle of attack.

\section{Conclusion}
This study, the first of its kind in the academic literature, highlights the significant impact of ceiling proximity effects on electroaerodynamic thruster performance. Taking advantage of ceiling proximity effects can increase the thrust and thrust efficiency of EAD-propelled MAVs by as much as $600\%$ due to the combination of fluid dynamic effects and the  electroadhesive force unique to EAD thrusters while in-ceiling-effect (ICE). We show that with a thin enough intake flange, this increase in efficiency comes with a net increase in thrust-to-weight ratio, improving prospects for future extended-duration power-autonomous flight.





\bibliographystyle{IEEEtran}

\bibliography{root}

\begin{thebibliography}{10}
\providecommand{\url}[1]{#1}
\csname url@rmstyle\endcsname
\providecommand{\newblock}{\relax}
\providecommand{\bibinfo}[2]{#2}
\providecommand\BIBentrySTDinterwordspacing{\spaceskip=0pt\relax}
\providecommand\BIBentryALTinterwordstretchfactor{4}
\providecommand\BIBentryALTinterwordspacing{\spaceskip=\fontdimen2\font plus
\BIBentryALTinterwordstretchfactor\fontdimen3\font minus \fontdimen4\font\relax}
\providecommand\BIBforeignlanguage[2]{{%
\expandafter\ifx\csname l@#1\endcsname\relax
\typeout{** WARNING: IEEEtran.bst: No hyphenation pattern has been}%
\typeout{** loaded for the language `#1'. Using the pattern for}%
\typeout{** the default language instead.}%
\else
\language=\csname l@#1\endcsname
\fi
#2}}

\bibitem{xu2023high}
J.~Xu, X.~Cai, S.~Cai, Y.~Shao, C.~Hu, S.~Lu, and S.~Ding, ``High-energy lithium-ion batteries: recent progress and a promising future in applications,'' \emph{Energy \& Environmental Materials}, vol.~6, no.~5, p. e12450, 2023.

\bibitem{mulgaonkar2014power}
Y.~Mulgaonkar, M.~Whitzer, B.~Morgan, C.~M. Kroninger, A.~M. Harrington, and V.~Kumar, ``Power and weight considerations in small, agile quadrotors,'' in \emph{Micro-and Nanotechnology Sensors, Systems, and Applications VI}, vol. 9083.\hskip 1em plus 0.5em minus 0.4em\relax SPIE, 2014, pp. 376--391.

\bibitem{raza2021energy}
W.~Raza, A.~Osman, F.~Ferrini, and F.~D. Natale, ``Energy-efficient inference on the edge exploiting tinyml capabilities for uavs,'' \emph{Drones}, vol.~5, no.~4, p. 127, 2021.

\bibitem{awasthi2020artificial}
S.~Awasthi, B.~Balusamy, and V.~Porkodi, ``Artificial intelligence supervised swarm uavs for reconnaissance,'' in \emph{Data Science and Analytics: 5th International Conference on Recent Developments in Science, Engineering and Technology, REDSET 2019, Gurugram, India, November 15--16, 2019, Revised Selected Papers, Part I 5}.\hskip 1em plus 0.5em minus 0.4em\relax Springer, 2020, pp. 375--388.

\bibitem{hafeez2023implementation}
A.~Hafeez, M.~A. Husain, S.~Singh, A.~Chauhan, M.~T. Khan, N.~Kumar, A.~Chauhan, and S.~Soni, ``Implementation of drone technology for farm monitoring \& pesticide spraying: A review,'' \emph{Information processing in Agriculture}, vol.~10, no.~2, pp. 192--203, 2023.

\bibitem{yoon2023cnn}
J.~Yoon, H.~Shin, K.~Kim, and S.~Lee, ``Cnn-and uav-based automatic 3d modeling methods for building exterior inspection,'' \emph{Buildings}, vol.~14, no.~1, p.~5, 2023.

\bibitem{elmokadem2021towards}
T.~Elmokadem and A.~V. Savkin, ``Towards fully autonomous uavs: A survey,'' \emph{Sensors}, vol.~21, no.~18, p. 6223, 2021.

\bibitem{ding2023aerodynamic}
R.~Ding, S.~Bai, K.~Dong, and P.~Chirarattananon, ``Aerodynamic effect for collision-free reactive navigation of a small quadcopter,'' \emph{npj Robotics}, vol.~1, no.~1, p.~2, 2023.

\bibitem{britcher2021use}
V.~Britcher and S.~Bergbreiter, ``Use of a mems differential pressure sensor to detect ground, ceiling, and walls on small quadrotors,'' \emph{IEEE Robotics and Automation Letters}, vol.~6, no.~3, pp. 4568--4575, 2021.

\bibitem{yu2023adaptive}
Y.~Yu, C.~Chen, J.~Guo, M.~Chadli, and Z.~Xiang, ``Adaptive formation control for unmanned aerial vehicles with collision avoidance and switching communication network,'' \emph{IEEE Transactions on Fuzzy Systems}, 2023.

\bibitem{liu2020adaptive}
S.~Liu, W.~Dong, Z.~Ma, and X.~Sheng, ``Adaptive aerial grasping and perching with dual elasticity combined suction cup,'' \emph{IEEE Robotics and Automation Letters}, vol.~5, no.~3, pp. 4766--4773, 2020.

\bibitem{thomas2013avian}
J.~Thomas, J.~Polin, K.~Sreenath, and V.~Kumar, ``Avian-inspired grasping for quadrotor micro uavs,'' in \emph{International Design Engineering Technical Conferences and Computers and Information in Engineering Conference}, vol. 55935.\hskip 1em plus 0.5em minus 0.4em\relax American Society of Mechanical Engineers, 2013, p. V06AT07A014.

\bibitem{hsiao2023energy}
Y.-H. Hsiao, S.~Bai, Y.~Zhou, H.~Jia, R.~Ding, Y.~Chen, Z.~Wang, and P.~Chirarattananon, ``Energy efficient perching and takeoff of a miniature rotorcraft,'' \emph{Communications Engineering}, vol.~2, no.~1, p.~38, 2023.

\bibitem{graule2016perching}
M.~A. Graule, P.~Chirarattananon, S.~B. Fuller, N.~T. Jafferis, K.~Y. Ma, M.~Spenko, R.~Kornbluh, and R.~J. Wood, ``Perching and takeoff of a robotic insect on overhangs using switchable electrostatic adhesion,'' \emph{Science}, vol. 352, no. 6288, pp. 978--982, 2016.

\bibitem{chen2019review}
C.~Chen and T.~Zhang, ``A review of design and fabrication of the bionic flapping wing micro air vehicles,'' \emph{Micromachines}, vol.~10, no.~2, p. 144, 2019.

\bibitem{drew2017first}
D.~S. Drew and K.~S. Pister, ``First takeoff of a flying microrobot with no moving parts,'' in \emph{2017 International Conference on Manipulation, Automation and Robotics at Small Scales (MARSS)}.\hskip 1em plus 0.5em minus 0.4em\relax IEEE, 2017.

\bibitem{drew2018toward}
D.~S. Drew, N.~O. Lambert, C.~B. Schindler, and K.~S. Pister, ``Toward controlled flight of the ionocraft: A flying microrobot using electrohydrodynamic thrust with onboard sensing and no moving parts,'' \emph{IEEE Robotics and Automation Letters}, vol.~3, no.~4, 2018.

\bibitem{zhang2022centimeter}
H.~Zhang, J.~Leng, D.~Liu, W.~Zhan, R.~Yun, Z.~Liu, M.~Qi, and X.~Yan, ``A centimeter-scale electrohydrodynamic multi-modal robot capable of rolling, hopping, and taking off,'' \emph{IEEE Robotics and Automation Letters}, vol.~7, no.~4, pp. 11\,791--11\,798, 2022.

\bibitem{nations2024empirical}
G.~Nations, C.~L. Nelson, and D.~S. Drew, ``Empirical study of ground proximity effects for small-scale electroaerodynamic thrusters,'' in \emph{2024 IEEE International Conference on Robotics and Automation (ICRA)}.\hskip 1em plus 0.5em minus 0.4em\relax IEEE, 2024, pp. 3868--3875.

\bibitem{monkman1997analysis}
G.~J. Monkman, ``An analysis of astrictive prehension,'' \emph{The International Journal of Robotics Research}, vol.~16, no.~1, pp. 1--10, 1997.

\bibitem{yamamoto2007wall}
A.~Yamamoto, T.~Nakashima, and T.~Higuchi, ``Wall climbing mechanisms using electrostatic attraction generated by flexible electrodes,'' in \emph{2007 International Symposium on Micro-NanoMechatronics and Human Science}.\hskip 1em plus 0.5em minus 0.4em\relax IEEE, 2007, pp. 389--394.

\bibitem{pekker2011model}
L.~Pekker and M.~Young, ``Model of ideal electrohydrodynamic thruster,'' \emph{journal of propulsion and power}, vol.~27, no.~4, 2011.

\bibitem{masuyama2013performance}
K.~Masuyama and S.~R. Barrett, ``On the performance of electrohydrodynamic propulsion,'' \emph{Proceedings of the Royal Society A: Mathematical, Physical and Engineering Sciences}, vol. 469, no. 2154, p. 20120623, 2013.

\bibitem{chen2013modeling}
R.~Chen, R.~Liu, and H.~Shen, ``Modeling and analysis of electric field and electrostatic adhesion force generated by interdigital electrodes for wall climbing robots,'' in \emph{2013 IEEE/RSJ International Conference on Intelligent Robots and Systems}.\hskip 1em plus 0.5em minus 0.4em\relax IEEE, 2013, pp. 2327--2332.

\bibitem{nelson2024high}
C.~L. Nelson and D.~S. Drew, ``High aspect ratio multi-stage ducted electroaerodynamic thrusters for micro air vehicle propulsion,'' \emph{IEEE Robotics and Automation Letters}, 2024.

\bibitem{hsiao2019ceiling}
Y.~H. Hsiao and P.~Chirarattananon, ``Ceiling effects for hybrid aerial--surface locomotion of small rotorcraft,'' \emph{IEEE/ASME Transactions on Mechatronics}, vol.~24, no.~5, pp. 2316--2327, 2019.

\bibitem{jimenez2019contact}
A.~E. Jimenez-Cano, P.~J. Sanchez-Cuevas, P.~Grau, A.~Ollero, and G.~Heredia, ``Contact-based bridge inspection multirotors: Design, modeling, and control considering the ceiling effect,'' \emph{IEEE Robotics and Automation Letters}, vol.~4, no.~4, pp. 3561--3568, 2019.

\bibitem{carter2021influence}
D.~J. Carter, L.~Bouchard, and D.~B. Quinn, ``Influence of the ground, ceiling, and sidewall on micro-quadrotors,'' \emph{AIAA Journal}, vol.~59, no.~4, pp. 1398--1405, 2021.

\bibitem{kocer2019aerial}
B.~B. Kocer, M.~E. Tiryaki, M.~Pratama, T.~Tjahjowidodo, and G.~G.~L. Seet, ``Aerial robot control in close proximity to ceiling: A force estimation-based nonlinear mpc,'' in \emph{2019 IEEE/RSJ International Conference on Intelligent Robots and Systems (IROS)}.\hskip 1em plus 0.5em minus 0.4em\relax IEEE, 2019, pp. 2813--2819.

\bibitem{hsiao2018ceiling}
Y.~H. Hsiao and P.~Chirarattananon, ``Ceiling effects for surface locomotion of small rotorcraft,'' in \emph{2018 IEEE/RSJ International Conference on Intelligent Robots and Systems (IROS)}.\hskip 1em plus 0.5em minus 0.4em\relax IEEE, 2018, pp. 6214--6219.

\bibitem{kocer2019uav}
B.~B. Kocer, V.~Kumtepeli, T.~Tjahjowidodo, M.~Pratama, A.~Tripathi, G.~S.~G. Lee, and Y.~Wang, ``Uav control in close proximities-ceiling effect on battery lifetime,'' in \emph{2019 2nd International Conference on Intelligent Autonomous Systems (ICoIAS)}.\hskip 1em plus 0.5em minus 0.4em\relax IEEE, 2019, pp. 193--197.

\bibitem{gao2019exploiting}
S.~Gao, C.~Di~Franco, D.~Carter, D.~Quinn, and N.~Bezzo, ``Exploiting ground and ceiling effects on autonomous uav motion planning,'' in \emph{2019 international conference on unmanned aircraft systems (ICUAS)}.\hskip 1em plus 0.5em minus 0.4em\relax IEEE, 2019, pp. 768--777.

\bibitem{du2024modelling}
I.~D. Du~Mutel, R.~Parin, D.~Carminati, and E.~Capello, ``Modelling of ground and ceiling effects for quadcopters based on experimental data,'' in \emph{2024 International Conference on Unmanned Aircraft Systems (ICUAS)}, 2024.

\bibitem{conyers2018empirical}
S.~A. Conyers, M.~J. Rutherford, and K.~P. Valavanis, ``An empirical evaluation of ceiling effect for small-scale rotorcraft,'' in \emph{2018 International Conference on Unmanned Aircraft Systems (ICUAS)}.\hskip 1em plus 0.5em minus 0.4em\relax IEEE, 2018, pp. 243--249.

\bibitem{jinjing2023aerodynamic}
H.~Jinjing, Y.~Zhang, Z.~Chao, C.~Songtao, and W.~Jianghao, ``Aerodynamic performance of hovering micro revolving wings in ground and ceiling effects at low reynolds number,'' \emph{Chinese Journal of Aeronautics}, vol.~36, no.~1, pp. 152--165, 2023.

\bibitem{liu2024analysis}
Y.~Liu, Z.~Kan, H.~Li, Y.~Gao, D.~Li, and S.~Zhao, ``Analysis and modeling of the aerodynamic ceiling effect on small-scale propellers with tilted angles,'' \emph{Aerospace Science and Technology}, vol. 147, p. 109038, 2024.

\bibitem{krape1968applications}
R.~P. Krape, ``Applications study of electroadhesive devices,'' NASA, Tech. Rep., 1968.

\bibitem{bamber2017visualization}
T.~Bamber, J.~Guo, J.~Singh, M.~Bigharaz, J.~Petzing, P.~A. Bingham, L.~Justham, J.~Penders, and M.~Jackson, ``Visualization methods for understanding the dynamic electroadhesion phenomenon,'' \emph{Journal of Physics D: Applied Physics}, vol.~50, no.~20, p. 205304, 2017.

\bibitem{tellez2011characterization}
J.~P.~D. Tellez, J.~Krahn, and C.~Menon, ``Characterization of electro-adhesives for robotic applications,'' in \emph{2011 IEEE international conference on robotics and biomimetics}.\hskip 1em plus 0.5em minus 0.4em\relax IEEE, 2011, pp. 1867--1872.

\bibitem{guo2019electroadhesion}
J.~Guo, J.~Leng, and J.~Rossiter, ``Electroadhesion technologies for robotics: A comprehensive review,'' \emph{IEEE Transactions on Robotics}, vol.~36, no.~2, pp. 313--327, 2019.

\bibitem{prahlad2008electroadhesive}
H.~Prahlad, R.~Pelrine, S.~Stanford, J.~Marlow, and R.~Kornbluh, ``Electroadhesive robots—wall climbing robots enabled by a novel, robust, and electrically controllable adhesion technology,'' in \emph{2008 IEEE international conference on robotics and automation}.\hskip 1em plus 0.5em minus 0.4em\relax IEEE, 2008, pp. 3028--3033.

\bibitem{shintake2016versatile}
J.~Shintake, S.~Rosset, B.~E. Schubert, D.~Floreano, and H.~Shea, ``Versatile soft grippers with intrinsic electroadhesion based on multifunctional polymer actuators,'' \emph{Advanced materials}, vol.~28, no.~2, pp. 231--238, 2016.

\bibitem{chen2022variable}
R.~Chen, Z.~Zhang, J.~Guo, F.~Liu, J.~Leng, and J.~Rossiter, ``Variable stiffness electroadhesion and compliant electroadhesive grippers,'' \emph{Soft Robotics}, vol.~9, no.~6, pp. 1074--1082, 2022.

\bibitem{hwang2021electroadhesion}
G.~Hwang, J.~Park, D.~S.~D. Cortes, K.~Hyeon, and K.-U. Kyung, ``Electroadhesion-based high-payload soft gripper with mechanically strengthened structure,'' \emph{IEEE Transactions on Industrial Electronics}, vol.~69, no.~1, pp. 642--651, 2021.

\bibitem{de2018inverted}
S.~D. De~Rivaz, B.~Goldberg, N.~Doshi, K.~Jayaram, J.~Zhou, and R.~J. Wood, ``Inverted and vertical climbing of a quadrupedal microrobot using electroadhesion,'' \emph{Science Robotics}, vol.~3, no.~25, 2018.

\bibitem{park2020lightweight}
S.~Park, D.~S. Drew, S.~Follmer, and J.~Rivas-Davila, ``Lightweight high voltage generator for untethered electroadhesive perching of micro air vehicles,'' \emph{IEEE Robotics and Automation Letters}, vol.~5, no.~3, pp. 4485--4492, 2020.

\bibitem{guo2015investigation}
J.~Guo, M.~Tailor, T.~Bamber, M.~Chamberlain, L.~Justham, and M.~Jackson, ``Investigation of relationship between interfacial electroadhesive force and surface texture,'' \emph{Journal of Physics D: Applied Physics}, vol.~49, no.~3, p. 035303, 2015.

\bibitem{guo2017experimental}
J.~Guo, T.~Bamber, J.~Petzing, L.~Justham, and M.~Jackson, ``Experimental study of relationship between interfacial electroadhesive force and applied voltage for different substrate materials,'' \emph{Applied Physics Letters}, vol. 110, no.~5, 2017.

\end{thebibliography}

\end{document}